\documentclass[wcp,x11names,table,xcdraw]{jmlr}
\usepackage{pgfplots}

\usepackage{bbm}
\usepackage{graphicx}
\usepackage{booktabs} 
\usepackage{amsmath,amsfonts}
\usepackage{multirow}
\usepackage{enumitem}
\usepackage{setspace}
\usepackage{tikz}
\usetikzlibrary{arrows.meta, positioning, calc}
\usepackage{pgfplots}
\pgfplotsset{compat=1.18}
\usepackage{caption}
\usepackage{subcaption}
\usepackage{colortbl} 
\makeatletter
\let\Ginclude@graphics\@org@Ginclude@graphics 
\makeatother

\jmlrvolume{304,}
\jmlryear{2025}
\jmlrworkshop{ACML 2025}

\title[KVCrush]{\textit{KVCrush:} \underline{K}ey \underline{V}alue \underline{C}ache size-\underline{r}eduction \underline{u}sing \underline{s}imilarity in \underline{h}ead-behaviour}

 \author{\Name{Gopi Krishna Jha} \Email{gopi.krishna.jha@intel.com}\\
 \Name{Sameh Gobriel} \Email{sameh.gobriel@intel.com}\\
 \Name{Liubov Talamanova} \Email{liubov.talamanova@intel.com}\\
  \Name{Nilesh Jain} \Email{nilesh.jain@intel.com}\\
\addr Intel Corporation}

\editors{Hung-yi Lee and Tongliang Liu}

\begin{document}

\makeatletter
\let \@jmlrpages \@empty
\makeatother

\maketitle

\begin{abstract}
Key-value (KV) caching has emerged as a crucial optimization technique for accelerating inference in large language models (LLMs). By allowing the attention operation to scale linearly rather than quadratically with the total sequence length, KV caching significantly enhances generation throughput~\citep{h2o}. However, due to large context lengths in the modern LLMs, the memory footprint of the KV is a huge bottleneck for model deployment directly impacting the model's batch size, hindering its ability to deliver high-throughput~\citep{scissorhands}. Existing research addresses this challenge using several techniques, such as discarding low-attention tokens, quantization, and matrix approximation which typically lead to a negative impact on the model accuracy. 

In this paper, we propose \emph{KVCrush} technology which can be combined with many KV compression technologies to improve the model accuracy at a much smaller memory. \emph{KVCrush} provides an alternate representation scheme for key-value states, along with a low-overhead token pruning algorithm that accounts for the token distribution in the KV cache, which in turn allows for a smaller footprint while maintaining the accuracy of the model. Based on our results, \emph{KVCrush} reduces \textit{LongBench}~\cite{longbench} KV Cache size by $4\times$ with less than $1\%$ accuracy drop and achieves state-of-the-art average accuracy with minimal overhead, incurring less than $0.5\%$ total inference latency. \emph{KVCrush} not only outperforms state-of-the-art importance-based token retention schemes in accuracy, but also integrates seamlessly with quantization, paging, and head-sharing techniques, requiring no retraining or architectural changes.

\end{abstract}


\section{Introduction}
Generative AI models, such as large language models (LLMs), have revolutionized the computing industry. These models boast an extensive number of parameters and consistently achieve state-of-the-art performance across various downstream tasks \cite{sun2019bert4rec} \cite{memrec} \cite{dosovitskiy2020image} \cite{raffel2020exploring}. However, the current trend of model size growth toward multi-trillion parameters~\cite{isaev2023scaling} — with models growing by one estimate~\cite{gholami2024ai} at a staggering rate of $410\times$ every 2 years— poses huge challenges for deploying them in practice. For instance, GPT-175B~\cite{gpt3} requires 325GB of memory just to load the model weights. Additionally, not only do inferences for these models strain the platform compute and memory resources (both in terms of bandwidth and capacity), but typically, this is also coupled with strict latency requirements (in the order of tens of milliseconds) which further complicates the problem and poses a significant engineering challenge for efficiently delivering high inference throughput while maintaining low latency for user requests.

LLMs typically consist of stacked transformer decoder layers, with the self-attention module being a critical building block. This module weighs the importance of different tokens—capturing their contextual relationships—by attending over key–value token pairs at different positions within the input sequence. Nevertheless, self-attention matrix multiplication is compute intensive and has a quadratic computational complexity with respect to sequence length \cite{allyouneed} which significantly impacts the inference throughput, especially for longer sequences \cite{survey}. 


KV caching has emerged as the de facto solution to this issue, converting the time complexity of token generation from quadratic to linear, albeit with increased memory overhead proportional to the context length. This mechanism is crucial for autoregressive tasks, where the model generates one output token at a time. In the decode phase, each token depends on the key and value tensors of all previous tokens (including the input tokens’ KV tensors computed at prefill and all new KV tensors computed until the current time step). To avoid recomputing these tensors for every token at each step, a KV cache stores them in memory, incrementally appending new entries to the cache as generation progresses.

\begin{equation}
\resizebox{0.60\textwidth}{!}{%
    $\text{KV Memory} = 2 \cdot B \cdot N_{L} \cdot N_{H} \cdot L_{\text{seq}} \cdot D \cdot precision$
}
\label{eq:kv_cache}
\end{equation}

The scalability of KV caching becomes a critical issue as models grow larger and more complex. The total memory used by a KV cache can be determined using Equation~\ref{eq:kv_cache}, where $B$ is the batch size, $N_{L}$ represents the number of layers in the model, $N_{H}$ represents the number of attention heads used, $D$ represents the dimensionality of the embeddings, $L_{seq}$ is the length of context in tokens, $precision$ is the number of bytes per value stored (e.g. 4B for FP32) and the factor $2$ is because two matrices for K and V are needed. As a result, for a given model, the KV cache size grows linearly with the maximum sequence length in the input context and the batch size, which, in practice, can result in an enormous KV cache size.

For instance, consider the OPT-175B model with its impressive 175 billion parameters, which consumes approximately 325 GB of memory. However, when using a batch size of 128 and a sequence length of only 8K, the KV cache requires around 4608 GB of memory. This is an order of magnitude (12X) larger than the model weights themselves. The issue of KV cache size has become increasingly prominent and is a significant cost factor in deploying large language models (LLMs). This is true, especially with the recent trend of LLM models that have been meticulously developed to scale up to handle extremely long context (for example, Google’s Gemini-pro-1.5 has shown to support a staggering 1M token context length~\cite{Gemini}  

Several approaches have been proposed to mitigate the memory bottleneck associated with KV caching. Recent research have explored different optimizations of KV caching, including approaches such as low-rank decomposition of the KV cache(e.g.,~\cite{decomp}) or pruning non-essential KV cache~\cite{h2o, snapkv, scissorhands, fastgen}, however, most of these techniques struggle to maintain the accuracy of the model at a smaller KV cache footprint when compared to the full model. 

To address this challenge, we introduce \emph{KVCrush}, a novel KV cache optimization that provides an alternate representation scheme for key-value states, along with a low-overhead token pruning algorithm that accounts for the token distribution in the KV cache. \emph{KVCrush}, in turn, allows for a smaller footprint while maintaining the accuracy of the model and can be easily combined with many KV cache compression technologies. It is a modular add-on that augments existing methods by preserving contextual diversity, and requires no model retraining or architectural changes. Furthermore, \emph{KVCrush} also works seamlessly with KV cache paging schemes (such as vLLM~\cite{vllm}) and mixed precision quantization~\cite{mixed} typically used in practical deployments. The technical contributions of this work can
be summarized as follows:
\begin{itemize}
    \item \textbf{Hardware-Efficient Alternative Representation of Tokens:} We leverage attention score patterns across heads to generate a binary feature vector for each token. This binary alternative representation is much smaller than the original key and value vectors, yet it preserves enough semantic information, to convey token importance and similarities, which we use to prune or retain the tokens very efficiently.
    \item \textbf{Accuracy-aware Low-Overhead KV Optimization Algorithm:} Our algorithm
    leverages the alternative binary representation of tokens and an \textit{anchor point} to bucketize tokens into representative groups, using very efficient low-overhead distance calculations that scale linearly with the number of tokens. This approach ensures that different token groups are represented in the reduced KV memory footprint, thereby maintaining the inference accuracy of the model.
    \item \textbf{Practical Efficiency and Integration:} We show that \emph{KVCrush} can be \textbf{efficiently combined} with token eviction-based KV cache compression methods. It achieves state-of-the-art average accuracy on \textit{LongBench}, with less than $0.5\%$ inference latency overhead, and offers up to $4\times$ cache reduction with under $1\%$ accuracy drop compared to the full cache.
\end{itemize}

\begin{figure*}[htp]
\centering
     \includegraphics[width=0.70\textwidth]{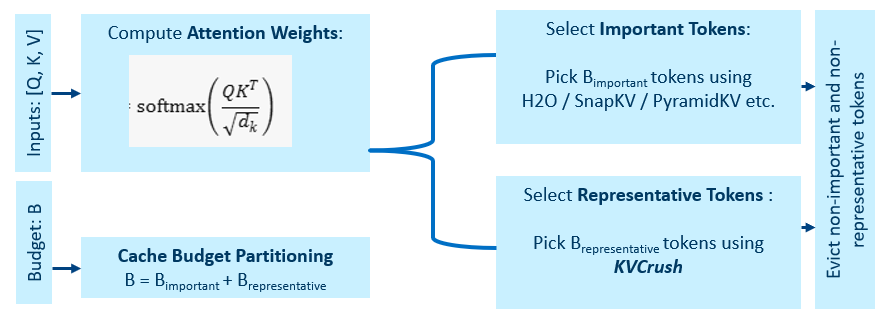}
     \caption{KVCrush flow: cache budget \(B\) is split into \(B_{important}\) (pivotal tokens via baseline methods) and \(B_{representative}\) (representative proxies selected using lightweight grouping).}
    \label{fig:kvcrush_flow}
\end{figure*}

\begin{algorithm}[ht]
\caption{Generate Binary Token Representation Using Per-Head Thresholding}
\label{algo:kvcrush_rep}
\SetAlgoLined
\KwIn{Query matrix $Q \in \mathbb{R}^{S \times D \times H}$, Key matrix $K \in \mathbb{R}^{S \times D \times H}$, thresholds $\theta_1, \ldots, \theta_H$}
\KwOut{Binary vector $b_t \in \{0,1\}^H$ for each token $t \in \{1, \ldots, S\}$}

Initialize $b_1, \ldots, b_S$ as zero vectors in $\{0,1\}^H$\;

\ForEach{head $h \in \{1, \ldots, H\}$}{
    Compute attention weights: $A_h = \text{softmax}\left(\frac{Q_h K_h^\top}{\sqrt{d_k}}\right) \in \mathbb{R}^{S \times S}$\;
    
    \ForEach{token $t \in \{1, \ldots, S\}$}{
        Compute normalized attention score: $w_h(t) = \frac{1}{S} \sum_{j=1}^{S} A_h(t, j)$\;
        
        \uIf{$w_h(t) \geq \theta_h$}{
            $b_t[h] \gets 1$\;
        }
        \Else{
            $b_t[h] \gets 0$\;
        }
    }
}
\Return $b_t$ for all tokens $t$
\end{algorithm}

\section{Related Work}
\label{background}

Techniques to reduce the size of KV Caches have received substantial research attention in recent years. Broadly speaking, we can divide these into three categories: 

\subsection{Quantizing KV Cache}
Quantization is the process of reducing the precision of a model’s parameters and activations to lower bit-widths to save memory and computational resources. Quantization can be categorized into post-training quantization (PTQ) and quantization-aware training (QAT), with PTQ often preferred for large language models due to its lower resource requirements. \cite{xiao2023smoothquant}, \cite{liu2024llmqat}, and \cite{sheng2023flexgen} demonstrated that quantizing queries, keys, and values to INT8 enables efficient attention operations. However, these methods apply quantization uniformly across all KV cache tokens, and thus will typically negatively impact the model generation accuracy.

On the other hand, research work such as \cite{mq1}, \cite{mq2}, \cite{mq3}, \cite{mikv} employ mixed precision quantization to allocate different number of bits to various model components or tensors, and thus enable a more efficient compression. This methods leverage the insight that different parts of a model exhibit varying levels of sensitivity to quantization. 

As we discuss later in detail, \emph{KVCrush} operates directly on the KV cache to determine which tokens to retain or evict. Since it does not alter the cache format itself, it is naturally complementary to quantization techniques. In fact, KVCrush can work seamlessly with mixed-precision KV caches by assigning different bitwidths to retained tokens based on their importance.

\subsection{Sharing KV Cache}
Multi-Query Attention (MQA)  and Grouped Query Attention (GQA)  are techniques developed to address the memory footprint issues LLMs allowing to share KV caches across heads. MQA, introduced by \citep{mqa}, reduces memory usage by sharing key and value representations across all attention heads, which enhances memory efficiency and inference speed, at the expense of generation quality. GQA, proposed by \citep{gqa}, extends this concept by grouping query heads to share KVs, balancing memory reduction with better performance retention. However, GQA involves higher training costs and complexity. Both methods offer trade-offs between memory efficiency and model performance, with MQA favoring memory savings and GQA providing a more balanced approach. 

Intuitively, \emph{KVCrush} is completely orthogonal to these KV cache sharing schemes. It operates directly on the deployed cache, regardless of whether key and value states are shared or grouped across attention heads.

\begin{algorithm}[t]
\caption{Token Grouping and Representative Selection via Hamming Clustering}
\label{algo:kvcrush_grouping}
\SetAlgoLined
\KwIn{Binary vectors $b_1, b_2, \ldots, b_S \in \{0,1\}^H$ for $S$ tokens, Number of buckets $B_{\text{representative}}$}
\KwOut{Selected representative tokens $R$}

Generate an anchor vector $a \in \{0,1\}^H$ using a chosen strategy (e.g., random, mean of input vectors, or alternating 0-1)\;

Initialize $B_{\text{representative}}$ empty buckets\;

\ForEach{token $b_t$}{
    Compute Hamming distance from $a$: $d_t = \text{Hamming}(b_t, a)$\;
    
    Assign $b_t$ to one of $B_{\text{representative}}$ buckets based on $d_t$\;
}

\ForEach{bucket}{
    Compute the centroid of all binary vectors in the bucket\;
    
    Select the token whose binary vector is closest to the centroid as representative\;
    
    Add representative to $R$\;
}

\Return $R$
\end{algorithm}

\subsection{Evicting inconsequential Keys and Values} \label{related:kv_eviction}
This category is the closest related work to \emph{KVCrush}. In this research category  different methods aim to prune key-value (KV) pairs from cache after input processing, aiming to enhance decoding efficiency. By evicting tokens out of KV cache, memory consumption is reduced, facilitating support for larger batch sizes and longer context windows. 

Different strategies for KV pruning and selectively dropping tokens from the KV cache have been proposed in recent research work, For example, in StreamLLM~\cite{streamingllm}, only the most recent tokens and attention sinks (first few tokens) are retained. H2O \cite{h2o} and Scissorhands \cite{scissorhands} utilize attention-based metrics to determine eviction, with H2O summing attention weights and Scissorhands considering the frequency of attention surpassing a threshold. FastGen \cite{fastgen} combines these methods with heuristics for special tokens. SnapKV~\cite{snapkv} similarly depends on attention weights to prune the KV cache, but addresses the efficiency during prefill with long context, and due to the complexity of computing attention score for the whole context, it limits the observation window to the final tokens of the input prompt, thus reducing the complexity from $O(L^2)$ to $O(L)$, where $L$ is the context length, while using max pooling to retain neighboring KVs. PyramidKV \cite{pyramidkv} builds on SnapKV by configuring variable eviction rates across layers, with more aggressive pruning in later layers where attention is less evenly distributed. DynamicKV \cite{dynamickv} adaptively resizes KV caches across layers based on task type. LaCache \cite{lacache} introduces a ladder-shaped caching pattern that prioritizes temporal coverage across layers.

The common drawback across these existing methods is that model accuracy drops as the KV compression ratio increases. These works focus solely on selecting a set of pivotal tokens (using varying algorithms) and evict the rest, which reduces KV cache size but at the cost of generation quality. \emph{KVCrush} addresses this by explicitly representing groups of evicted tokens via a small set of representative tokens, preserving context diversity and mitigating accuracy loss under high compression. It ensures that different token groups are reflected in the reduced KV memory footprint, thereby sustaining accuracy. As a result, \emph{KVCrush} can be combined with many existing methods to remedy accuracy drop at the same KV cache budget. We present results in Section~\ref{experiments}, combining \emph{KVCrush} with state-of-the-art techniques and demonstrating improved performance at equal compression ratios.


        \begin{table*}[ht!]
        \resizebox{\textwidth}{!}
        {
            \begin{tabular}{@{}ccc@{}}
            \toprule
            \textbf{Benchmarks}                  & \textbf{LongBench}             & \textbf{lm-eval-harness}                  \\ \midrule
            \textbf{Experiments}                 & Section ~\ref{sec:end_to_end}, Section ~\ref{sec:cache_budget}                       & Section ~\ref{sec:anchor_points}, Section ~\ref{sec:kmeans_comparison}, Section ~\ref{sec:kvcrush_int}, Section ~\ref{sec:kvcrush_paged}                             \\
            \textbf{Datasets} &
              \begin{tabular}[c]{@{}c@{}}narrativeqa, qasper,   multifieldqa\_en, hotpotqa, 2wikimqa, \\      musique, gov\_report, qmsum, multi\_news, trec, triviaqa, \\      samsum,  passage\_count,   passage\_retrieval\_en,   repobench-p\end{tabular} &
              GSM8K and XSUM \\
            \textbf{Models} &
              Mistral-7B-Instruct-v0.2, Meta-Llama-3-8B-Instruct &
              Phi-3-mini-4k-instruct, Meta-Llama-3-8B-Instruct,  Llama-2-7b-chat-hf \\
            \textbf{Baselines}                   & FullKV, H2O, SnapKV, PyramidKV & H2O                                       \\
            \textbf{Paging Mode}                 & Token Level, Chunk Level       & Page Level (Page Size = 32)                               \\
            \textbf{Total Cache Budget (tokens)} & 2048                           & 672 = 32(initial)+512(middle)+128(recent) \\
            \textbf{Cache Budget Partitioning}   & 25\% KVCrush + 75\% Baseline   & 25\% KVCrush + 75\% Baseline   \\
            \textbf{Cuda Version}                & 12.2                           & 12.2                                      \\
            \textbf{Pytorch Version}             & 2.4.1+cu121                  & 2.4.1+cu121                             \\ \bottomrule
            \end{tabular}
        }
        \caption{Experimental settings used for evaluation}
        \label{table:settings}
        \end{table*}

        \begin{table*}[ht!]
        \resizebox{\textwidth}{!}
        {
        \begin{tabular}{|c|c|cc|cc|cc|}
        \hline
                                   & Cache Budget: 512 & \multicolumn{2}{c|}{\textbf{Phi-3-mini-4k-instruct}}                                                                           & \multicolumn{2}{c|}{\textbf{Meta-Llama-3-8B-Instruct}}                                                                         & \multicolumn{2}{c|}{\textbf{Llama-2-7b-chat-hf}}                                 \\ \hline
                                   & hh-cl             & \multicolumn{1}{c|}{\textbf{Strict}}                                     & \textbf{Flexible}                                   & \multicolumn{1}{c|}{\textbf{Strict}}                                     & \textbf{Flexible}                                   & \multicolumn{1}{c|}{\textbf{Strict}}              & \textbf{Flexible}            \\ \hline
        \textit{H2O}               & 512-0             & \multicolumn{1}{c|}{70.7}                                                & 79.3                                                & \multicolumn{1}{c|}{74.9}                                                & 74.7                                                & \multicolumn{1}{c|}{0.209}                        & 0.225                        \\ \hline
        \textit{kvcrush.random}    & 128-384           & \multicolumn{1}{c|}{{\color[HTML]{00B050} 75.4}}                         & {\color[HTML]{00B050} 80.9}                         & \multicolumn{1}{c|}{{\color[HTML]{00B050} 76.2}}                         & {\color[HTML]{00B050} 76.2}                         & \multicolumn{1}{c|}{{\color[HTML]{00B050} 0.211}} & {\color[HTML]{00B050} 0.226} \\ \hline
        \textit{kvcrush.mean}      & 128-384           & \multicolumn{1}{c|}{\cellcolor[HTML]{FFFFFF}{\color[HTML]{00B050} 75.2}} & \cellcolor[HTML]{FFFFFF}{\color[HTML]{00B050} 80.6} & \multicolumn{1}{c|}{\cellcolor[HTML]{FFFFFF}{\color[HTML]{00B050} 76.5}} & \cellcolor[HTML]{FFFFFF}{\color[HTML]{00B050} 76.4} & \multicolumn{1}{c|}{{\color[HTML]{00B050} 0.21}}  & {\color[HTML]{00B050} 0.229} \\ \hline
        \textit{kvcrush.alternate} & 128-384           & \multicolumn{1}{c|}{\cellcolor[HTML]{FFFFFF}{\color[HTML]{00B050} 74.6}} & \cellcolor[HTML]{FFFFFF}{\color[HTML]{00B050} 80.9} & \multicolumn{1}{c|}{\cellcolor[HTML]{FFFFFF}{\color[HTML]{00B050} 75.7}} & \cellcolor[HTML]{FFFFFF}{\color[HTML]{00B050} 75.6} & \multicolumn{1}{c|}{{\color[HTML]{00B050} 0.212}} & {\color[HTML]{00B050} 0.227} \\ \hline
        \end{tabular}
        }
        \caption{GSM-8K Accuracy using different anchor points in KVCrush. KVCrush outperforms the baseline H2O even using generic anchor points like random, mean and alternate 0s and 1s. Here hh and cl represents the cache budget used by H2O and KVCrush respectively.}
        \label{table:anchor}
        \end{table*}

\section{KVCrush}
\label{sec:kvcrush}
The basic flow of KVCrush is shown in Figure~\ref{fig:kvcrush_flow}. Given a certain total budget $B$ for the KV Cache, this is split into two smaller portions, $B_{important}$ represents the cache budget available to store the set of pivotal tokens. This set is determined based on the specifics of KV cache compression algorithm used (e.g. H20, SnapKV, PyramidKV, etc.). While $B_{representative}$ represents the cache budget available to store along a set of representative tokens, these act as proxies of the evicted tokens, and are selected based on low-overhead grouping algorithm (discussed in Section~\ref{kvcrush.grouping}) to ensure better representation of the whole context.   

\subsection{KVCrush Alternative Token Representation} \label{kvcrush.rep}
In the KV cache tokens are represented as floating-point vectors of size $D$ each. $D$ is the size of embedding length and is typically not small. For example, for GPT-3~\cite{gpt3} D is 2048 dimensions while for LLaMA-65B~\cite{llama} it is set at 4096 dimensions. As previously mentioned, KVCrush will try to group tokens in order to select representative tokens from each group. As a result, it is essential to minimize the overhead of the grouping algorithm. The running time of any clustering algorithm in $D$ dimensions will be proportional to the value of $D$.  

To minimize this overhead, KVCrush tries to have an alternative representation of each token with a vector of length $<< D$. The main challenge is how to find this smaller  representation and yet still preserve sufficient semantic information of the tokens to be able to differentiate and group them based on their contextual relationship.

KVCrush draws some insights of alternative token representation from the FastGen research~\cite{fastgen}. The authors in this work demonstrated that distinct attention heads exhibit unique structural patterns and attend to different tokens differently. For example, some attention heads will attend more to special tokens, others will attend more to locality among tokens, others will attend more to punctuation, etc.

Building on these findings we can deduce that the attention score of a certain token across $H$ heads is a vector of length $H$ and will represent good semantic information about the contextual properties of that token. Keeping in mind that $H<<D$, where for example, $H$ is 96, 128 for GPT-3~\cite{gpt3} and LLaMA-65B~\cite{llama}, respectively.       

To reduce the grouping overhead even more, KVCrush takes the alternative token representation one step further. Instead of using a floating point vector of size $H$ each, it converts that into a binary vector of size $H$ replacing the floating point with a single bit. The main insight is that, for a given head and attention score based token eviction algorithm, the binary decision of whether to retain or evict a token encodes important contextual signals (e.g., attention score, recency), which we use to build token-level semantic fingerprints.

Algorithm~\ref{algo:kvcrush_rep} depicts how \emph{KVCrush} generates a binary representation of size $H$ for each input token. The process can be summarized in the following steps:

\begin{itemize}
    \item \textbf{Compute the attention weight matrix:} For each head $h$, compute the attention weight matrix $A_h \in \mathbb{R}^{S \times S}$ using the query and key projections. Note that this attention computation is already performed during inference, and \emph{KVCrush} leverages these values without recomputation.
    
    \item \textbf{Apply a per-head threshold:} For each token $t$ and head $h$, compute the row-wise normalized attention score $w_h(t) = \frac{1}{S} \sum_{j=1}^{S} A_h(t, j)$. A token is marked with a bit value of 1 if $w_h(t) \geq \theta_h$, and 0 otherwise. The threshold $\theta_h$ is chosen such that it retains a target fraction of tokens for each head, based on desired compression ratio.
    
    \item \textbf{Form the binary representation:} For each token $t$, collate the $H$ bit values across all heads into a binary vector $b_t \in \{0,1\}^H$, which captures its cross-head importance signature.
\end{itemize}

\subsection{KVCrush Token Grouping and Cache Eviction} \label{kvcrush.grouping}
 
 As previously mentioned, KVCrush will select $B_{representative}$ tokens that act as proxies of the evicted tokens based to keep in the KV Cache. The selection is based on a low-overhead grouping algorithm built on top of the alternative binary representation of tokens. Algorithm~\ref{algo:kvcrush_grouping} depicts the main steps of this weak clustering algorithm to form token groups and can be summarized as follows:  

\begin{itemize}
\item The clustering algorithm takes a set of $S$ tokens each represented with a binary vector of length $H$ as input, it selects $B_{representative}$ tokens where $B_{representative} < S$ vectors as output
\item	An arbitrary anchor point is first selected (ideally, the anchor point should be centrally placed to all $S$ vectors in an $H$ dimension space). In our experiments we present results for 3 different (low-overhead) selection of anchor points (random, alternate and mean). 
\item	For each of the $S$ input vectors, hamming distance is computed with respect to the anchor point. The binary representation enables computing Hamming distances on-the-fly using lightweight bitwise operations, making the grouping process extremely efficient compared to standard clustering.

\item	Each token is then assigned to one of the $B_{representative}$ buckets. The selected bucket is the one with the lowest hamming distance.   
\item	After the bucketization of all \(S\) tokens into their corresponding buckets, one representative vector \emph{nearest to the centroid} of each bucket is selected, and the remaining \((S - B_{representative})\) tokens are dropped. The retained tokens, along with the \(B_{important}\) tokens preserved by the KV compression algorithm, form the final \(B\) tokens of the KV cache.

\item After assigning all $S$ tokens to their respective buckets, one representative vector is selected from each by finding the token whose binary representation is closest to that bucket’s centroid. All the other ($S - B_{\text{representative}}$) tokens are dropped. Those retained tokens, in addition to the $B_{\text{important}}$ tokens retained by the KV compression algorithm, represent the final $B$ tokens of the KV cache.

 \end{itemize}

 It should be mentioned however, that we described KVCrush grouping algorithm using Hamming distance computations with respect to ONE anchor point, and thus, token pruning only requires $S$ Hamming distance comparisons instead of using $O(S^2)$ distance computations required by standard clustering algorithms. In Section~\ref{sec:kmeans_comparison} we present results for using a higher overhead clustering algorithm on accuracy and latency of \emph{KVCrush}.

\section{Experiments}
\label{experiments}

We organize our experiments to evaluate \emph{KVCrush} along four dimensions: accuracy and latency trade-offs under constrained cache budgets; ablation studies on anchor selection, budget partitioning, and clustering methods (including comparisons with k-means); integration with existing KV eviction methods such as H2O, SnapKV, and PyramidKV; and performance in paged KV cache settings.

\subsection{Experimental Setup}
We summarize experimental setup used for Evaluation in table~\ref{table:settings}.
\subsubsection{Baselines}
  For baseline comparison, we evaluate \emph{KVCrush} alongside four methods: \textbf{FullKV} uses the complete KV cache with zero eviction. \textbf{H2O} \citep{h2o} determines eviction based on the cumulative attention weights, retaining the top-$n$ tokens per layer. \textbf{SnapKV} \citep{snapkv} restricts the observation window to the final tokens of the prompt while employing max pooling to preserve neighboring KVs. \textbf{PyramidKV} \citep{pyramidkv} extends SnapKV by applying variable eviction rates across layers, pruning more aggressively in later layers where attention distributions are less uniform.

\subsubsection{Datasets and Models}
We evaluate \emph{KVCrush} on 16 diverse \textit{LongBench} datasets and the GSM-8K and XSUM tasks from \textit{lm-eval-harness}, using \textit{LLaMa-3-8B-Instruct}, \textit{Mistral-7B-Instruct-v0.2}, and \textit{Phi-3-mini-4k-instruct} models. Details of datasets, models, and input lengths are summarized in Table~\ref{table:settings}.

\subsection{End-to-End Evaluation}
\label{sec:end_to_end}
\subsubsection{Accuracy}
\label{sec:results_accuracy}

\begin{table}[t]

\resizebox{\textwidth}{!}{
\begin{tabular}{|c|cccccccccccccccc|}
\hline
 &
  \multicolumn{3}{c|}{\textbf{Single Document QA}} &
  \multicolumn{3}{c|}{\textbf{Multi Document QA}} &
  \multicolumn{3}{c|}{\textbf{Summarization}} &
  \multicolumn{3}{c|}{\textbf{Few Shot Learning}} &
  \multicolumn{2}{c|}{\textbf{Synthetic}} &
  \multicolumn{1}{c|}{\textbf{Code}} &
  \textbf{} \\ \hline
 &
  \multicolumn{1}{c|}{\rotatebox[origin=c]{90}{narrativeqa}} &
  \multicolumn{1}{c|}{\rotatebox[origin=c]{90}{qasper}} &
  \multicolumn{1}{c|}{\rotatebox[origin=c]{90}{multifieldqa\_en}} &
  \multicolumn{1}{c|}{\rotatebox[origin=c]{90}{hotpotqa}} &
  \multicolumn{1}{c|}{\rotatebox[origin=c]{90}{2wikimqa}} &
  \multicolumn{1}{c|}{\rotatebox[origin=c]{90}{musique}} &
  \multicolumn{1}{c|}{\rotatebox[origin=c]{90}{gov\_report}} &
  \multicolumn{1}{c|}{\rotatebox[origin=c]{90}{qmsum}} &
  \multicolumn{1}{c|}{\rotatebox[origin=c]{90}{multi\_news}} &
  \multicolumn{1}{c|}{\rotatebox[origin=c]{90}{trec}} &
  \multicolumn{1}{c|}{\rotatebox[origin=c]{90}{triviaqa}} &
  \multicolumn{1}{c|}{\rotatebox[origin=c]{90}{samsum}} &
  \multicolumn{1}{c|}{\rotatebox[origin=c]{90}{psg\_count}} &
  \multicolumn{1}{c|}{\rotatebox[origin=c]{90}{psg\_ret\_en}} &
  \multicolumn{1}{c|}{\rotatebox[origin=c]{90}{repobench-p}} &
  \rotatebox[origin=c]{90}{Average} \\ \hline
 &
  \multicolumn{1}{c|}{18409} &
  \multicolumn{1}{c|}{3619} &
  \multicolumn{1}{c|}{4559} &
  \multicolumn{1}{c|}{9151} &
  \multicolumn{1}{c|}{4887} &
  \multicolumn{1}{c|}{11214} &
  \multicolumn{1}{c|}{8734} &
  \multicolumn{1}{c|}{10614} &
  \multicolumn{1}{c|}{2113} &
  \multicolumn{1}{c|}{5177} &
  \multicolumn{1}{c|}{8209} &
  \multicolumn{1}{c|}{6258} &
  \multicolumn{1}{c|}{11141} &
  \multicolumn{1}{c|}{9289} &
  \multicolumn{1}{c|}{4206} &
  7839 \\ \hline
  
Compression   (x) &
  \multicolumn{1}{c|}{9.0} &
  \multicolumn{1}{c|}{1.8} &
  \multicolumn{1}{c|}{2.2} &
  \multicolumn{1}{c|}{4.5} &
  \multicolumn{1}{c|}{2.4} &
  \multicolumn{1}{c|}{5.5} &
  \multicolumn{1}{c|}{4.3} &
  \multicolumn{1}{c|}{5.2} &
  \multicolumn{1}{c|}{1.0} &
  \multicolumn{1}{c|}{2.5} &
  \multicolumn{1}{c|}{4.0} &
  \multicolumn{1}{c|}{3.1} &
  \multicolumn{1}{c|}{5.4} &
  \multicolumn{1}{c|}{4.5} &
  \multicolumn{1}{c|}{2.1} &
  4 \\ \hline
 &
  \multicolumn{16}{c|}{\textbf{Mistral-7B-Instruct-v0.2   (Cache Budget: 2048)}} \\ \hline
FullKV &
  \multicolumn{1}{c|}{26.85} &
  \multicolumn{1}{c|}{33.06} &
  \multicolumn{1}{c|}{49.44} &
  \multicolumn{1}{c|}{43.02} &
  \multicolumn{1}{c|}{27.33} &
  \multicolumn{1}{c|}{18.78} &
  \multicolumn{1}{c|}{32.87} &
  \multicolumn{1}{c|}{24.21} &
  \multicolumn{1}{c|}{27.05} &
  \multicolumn{1}{c|}{71.00} &
  \multicolumn{1}{c|}{86.23} &
  \multicolumn{1}{c|}{42.90} &
  \multicolumn{1}{c|}{2.75} &
  \multicolumn{1}{c|}{86.98} &
  \multicolumn{1}{c|}{54.41} &
  41.79 \\ \hline
\rowcolor[HTML]{FFFC9E} 
H2O &
  \multicolumn{1}{c|}{\cellcolor[HTML]{FFFC9E}25.10} &
  \multicolumn{1}{c|}{\cellcolor[HTML]{FFFC9E}30.67} &
  \multicolumn{1}{c|}{\cellcolor[HTML]{FFFC9E}48.29} &
  \multicolumn{1}{c|}{\cellcolor[HTML]{FFFC9E}40.89} &
  \multicolumn{1}{c|}{\cellcolor[HTML]{FFFC9E}25.98} &
  \multicolumn{1}{c|}{\cellcolor[HTML]{FFFC9E}15.57} &
  \multicolumn{1}{c|}{\cellcolor[HTML]{FFFC9E}28.06} &
  \multicolumn{1}{c|}{\cellcolor[HTML]{FFFC9E}23.48} &
  \multicolumn{1}{c|}{\cellcolor[HTML]{FFFC9E}26.78} &
  \multicolumn{1}{c|}{\cellcolor[HTML]{FFFC9E}\textbf{60.50}} &
  \multicolumn{1}{c|}{\cellcolor[HTML]{FFFC9E}{\color[HTML]{00B050} \textbf{86.33}}} &
  \multicolumn{1}{c|}{\cellcolor[HTML]{FFFC9E}42.48} &
  \multicolumn{1}{c|}{\cellcolor[HTML]{FFFC9E}2.57} &
  \multicolumn{1}{c|}{\cellcolor[HTML]{FFFC9E}82.73} &
  \multicolumn{1}{c|}{\cellcolor[HTML]{FFFC9E}52.92} &
  39.49 \\ \hline
\rowcolor[HTML]{FFFC9E} 
H2O+KVCrush &
  \multicolumn{1}{c|}{\cellcolor[HTML]{FFFC9E}\textbf{25.77}} &
  \multicolumn{1}{c|}{\cellcolor[HTML]{FFFC9E}\textbf{30.78}} &
  \multicolumn{1}{c|}{\cellcolor[HTML]{FFFC9E}\textbf{48.47}} &
  \multicolumn{1}{c|}{\cellcolor[HTML]{FFFC9E}\textbf{41.29}} &
  \multicolumn{1}{c|}{\cellcolor[HTML]{FFFC9E}\textbf{26.33}} &
  \multicolumn{1}{c|}{\cellcolor[HTML]{FFFC9E}\textbf{16.42}} &
  \multicolumn{1}{c|}{\cellcolor[HTML]{FFFC9E}\textbf{28.87}} &
  \multicolumn{1}{c|}{\cellcolor[HTML]{FFFC9E}\textbf{23.62}} &
  \multicolumn{1}{c|}{\cellcolor[HTML]{FFFC9E}\textbf{26.83}} &
  \multicolumn{1}{c|}{\cellcolor[HTML]{FFFC9E}65.42} &
  \multicolumn{1}{c|}{\cellcolor[HTML]{FFFC9E}86.26} &
  \multicolumn{1}{c|}{\cellcolor[HTML]{FFFC9E}\textbf{42.93}} &
  \multicolumn{1}{c|}{\cellcolor[HTML]{FFFC9E}\textbf{2.67}} &
  \multicolumn{1}{c|}{\cellcolor[HTML]{FFFC9E}\textbf{83.44}} &
  \multicolumn{1}{c|}{\cellcolor[HTML]{FFFC9E}\textbf{53.21}} &
  \textbf{40.15} \\ \hline
\rowcolor[HTML]{CBCEFB} 
SnapKV &
  \multicolumn{1}{c|}{\cellcolor[HTML]{CBCEFB}26.15} &
  \multicolumn{1}{c|}{\cellcolor[HTML]{CBCEFB}{\color[HTML]{00B050} \textbf{32.38}}} &
  \multicolumn{1}{c|}{\cellcolor[HTML]{CBCEFB}49.54} &
  \multicolumn{1}{c|}{\cellcolor[HTML]{CBCEFB}41.66} &
  \multicolumn{1}{c|}{\cellcolor[HTML]{CBCEFB}27.54} &
  \multicolumn{1}{c|}{\cellcolor[HTML]{CBCEFB}{\color[HTML]{00B050} \textbf{19.43}}} &
  \multicolumn{1}{c|}{\cellcolor[HTML]{CBCEFB}29.58} &
  \multicolumn{1}{c|}{\cellcolor[HTML]{CBCEFB}23.83} &
  \multicolumn{1}{c|}{\cellcolor[HTML]{CBCEFB}26.70} &
  \multicolumn{1}{c|}{\cellcolor[HTML]{CBCEFB}71.00} &
  \multicolumn{1}{c|}{\cellcolor[HTML]{CBCEFB}86.31} &
  \multicolumn{1}{c|}{\cellcolor[HTML]{CBCEFB}\textbf{43.07}} &
  \multicolumn{1}{c|}{\cellcolor[HTML]{CBCEFB}{\color[HTML]{00B050} \textbf{2.89}}} &
  \multicolumn{1}{c|}{\cellcolor[HTML]{CBCEFB}85.89} &
  \multicolumn{1}{c|}{\cellcolor[HTML]{CBCEFB}53.87} &
  41.32 \\ \hline
\rowcolor[HTML]{CBCEFB} 
SnapKV+KVCrush &
  \multicolumn{1}{c|}{\cellcolor[HTML]{CBCEFB}\textbf{26.20}} &
  \multicolumn{1}{c|}{\cellcolor[HTML]{CBCEFB}\textbf{32.09}} &
  \multicolumn{1}{c|}{\cellcolor[HTML]{CBCEFB}\textbf{49.86}} &
  \multicolumn{1}{c|}{\cellcolor[HTML]{CBCEFB}\textbf{41.86}} &
  \multicolumn{1}{c|}{\cellcolor[HTML]{CBCEFB}\textbf{28.33}} &
  \multicolumn{1}{c|}{\cellcolor[HTML]{CBCEFB}18.85} &
  \multicolumn{1}{c|}{\cellcolor[HTML]{CBCEFB}\textbf{29.62}} &
  \multicolumn{1}{c|}{\cellcolor[HTML]{CBCEFB}\textbf{23.92}} &
  \multicolumn{1}{c|}{\cellcolor[HTML]{CBCEFB}\textbf{26.95}} &
  \multicolumn{1}{c|}{\cellcolor[HTML]{CBCEFB}\textbf{71.00}} &
  \multicolumn{1}{c|}{\cellcolor[HTML]{CBCEFB}\textbf{86.20}} &
  \multicolumn{1}{c|}{\cellcolor[HTML]{CBCEFB}42.99} &
  \multicolumn{1}{c|}{\cellcolor[HTML]{CBCEFB}2.88} &
  \multicolumn{1}{c|}{\cellcolor[HTML]{CBCEFB}\textbf{86.06}} &
  \multicolumn{1}{c|}{\cellcolor[HTML]{CBCEFB}\textbf{54.13}} &
  \textbf{41.40} \\ \hline
\rowcolor[HTML]{38FFF8} 
PyramidKV &
  \multicolumn{1}{c|}{\cellcolor[HTML]{38FFF8}25.82} &
  \multicolumn{1}{c|}{\cellcolor[HTML]{38FFF8}31.67} &
  \multicolumn{1}{c|}{\cellcolor[HTML]{38FFF8}49.20} &
  \multicolumn{1}{c|}{\cellcolor[HTML]{38FFF8}41.19} &
  \multicolumn{1}{c|}{\cellcolor[HTML]{38FFF8}27.01} &
  \multicolumn{1}{c|}{\cellcolor[HTML]{38FFF8}\textbf{19.37}} &
  \multicolumn{1}{c|}{\cellcolor[HTML]{38FFF8}29.15} &
  \multicolumn{1}{c|}{\cellcolor[HTML]{38FFF8}23.89} &
  \multicolumn{1}{c|}{\cellcolor[HTML]{38FFF8}26.72} &
  \multicolumn{1}{c|}{\cellcolor[HTML]{38FFF8}71.00} &
  \multicolumn{1}{c|}{\cellcolor[HTML]{38FFF8}86.28} &
  \multicolumn{1}{c|}{\cellcolor[HTML]{38FFF8}{\color[HTML]{00B050} \textbf{43.24}}} &
  \multicolumn{1}{c|}{\cellcolor[HTML]{38FFF8}2.73} &
  \multicolumn{1}{c|}{\cellcolor[HTML]{38FFF8}85.06} &
  \multicolumn{1}{c|}{\cellcolor[HTML]{38FFF8}53.57} &
  41.06 \\ \hline
\rowcolor[HTML]{38FFF8} 
PyramidKV+KVCrush &
  \multicolumn{1}{c|}{\cellcolor[HTML]{38FFF8}\textbf{26.05}} &
  \multicolumn{1}{c|}{\cellcolor[HTML]{38FFF8}\textbf{31.99}} &
  \multicolumn{1}{c|}{\cellcolor[HTML]{38FFF8}\textbf{49.30}} &
  \multicolumn{1}{c|}{\cellcolor[HTML]{38FFF8}\textbf{41.37}} &
  \multicolumn{1}{c|}{\cellcolor[HTML]{38FFF8}\textbf{27.12}} &
  \multicolumn{1}{c|}{\cellcolor[HTML]{38FFF8}18.79} &
  \multicolumn{1}{c|}{\cellcolor[HTML]{38FFF8}\textbf{29.15}} &
  \multicolumn{1}{c|}{\cellcolor[HTML]{38FFF8}\textbf{24.22}} &
  \multicolumn{1}{c|}{\cellcolor[HTML]{38FFF8}\textbf{27.05}} &
  \multicolumn{1}{c|}{\cellcolor[HTML]{38FFF8}\textbf{71.00}} &
  \multicolumn{1}{c|}{\cellcolor[HTML]{38FFF8}\textbf{86.25}} &
  \multicolumn{1}{c|}{\cellcolor[HTML]{38FFF8}43.01} &
  \multicolumn{1}{c|}{\cellcolor[HTML]{38FFF8}\textbf{2.77}} &
  \multicolumn{1}{c|}{\cellcolor[HTML]{38FFF8}\textbf{85.23}} &
  \multicolumn{1}{c|}{\cellcolor[HTML]{38FFF8}\textbf{53.87}} &
  \textbf{41.14} \\ \hline
KVCrush* &
  \multicolumn{1}{c|}{{\color[HTML]{00B050} \textbf{26.20}}} &
  \multicolumn{1}{c|}{32.09} &
  \multicolumn{1}{c|}{{\color[HTML]{00B050} \textbf{49.86}}} &
  \multicolumn{1}{c|}{{\color[HTML]{00B050} \textbf{41.86}}} &
  \multicolumn{1}{c|}{{\color[HTML]{00B050} \textbf{28.33}}} &
  \multicolumn{1}{c|}{18.85} &
  \multicolumn{1}{c|}{{\color[HTML]{00B050} \textbf{29.62}}} &
  \multicolumn{1}{c|}{{\color[HTML]{00B050} \textbf{24.22}}} &
  \multicolumn{1}{c|}{{\color[HTML]{00B050} \textbf{27.05}}} &
  \multicolumn{1}{c|}{{\color[HTML]{00B050} \textbf{71.00}}} &
  \multicolumn{1}{c|}{86.26} &
  \multicolumn{1}{c|}{43.01} &
  \multicolumn{1}{c|}{2.88} &
  \multicolumn{1}{c|}{{\color[HTML]{00B050} \textbf{86.06}}} &
  \multicolumn{1}{c|}{{\color[HTML]{00B050} \textbf{54.13}}} &
  {\color[HTML]{00B050} \textbf{41.43}} \\ \hline
 &
  \multicolumn{16}{c|}{\textbf{LLaMa-3-8B-Instruct  (Cache Budget: 2048)}} \\ \hline
FullKV &
  \multicolumn{1}{c|}{\cellcolor[HTML]{FFFFFF}25.56} &
  \multicolumn{1}{c|}{\cellcolor[HTML]{FFFFFF}31.95} &
  \multicolumn{1}{c|}{\cellcolor[HTML]{FFFFFF}39.71} &
  \multicolumn{1}{c|}{\cellcolor[HTML]{FFFFFF}43.56} &
  \multicolumn{1}{c|}{\cellcolor[HTML]{FFFFFF}35.63} &
  \multicolumn{1}{c|}{\cellcolor[HTML]{FFFFFF}21.18} &
  \multicolumn{1}{c|}{\cellcolor[HTML]{FFFFFF}28.58} &
  \multicolumn{1}{c|}{\cellcolor[HTML]{FFFFFF}23.27} &
  \multicolumn{1}{c|}{\cellcolor[HTML]{FFFFFF}26.75} &
  \multicolumn{1}{c|}{\cellcolor[HTML]{FFFFFF}74.00} &
  \multicolumn{1}{c|}{\cellcolor[HTML]{FFFFFF}90.48} &
  \multicolumn{1}{c|}{\cellcolor[HTML]{FFFFFF}42.30} &
  \multicolumn{1}{c|}{\cellcolor[HTML]{FFFFFF}4.80} &
  \multicolumn{1}{c|}{\cellcolor[HTML]{FFFFFF}69.25} &
  \multicolumn{1}{c|}{\cellcolor[HTML]{FFFFFF}53.92} &
  40.73 \\ \hline
\rowcolor[HTML]{FFFC9E} 
H2O &
  \multicolumn{1}{c|}{\cellcolor[HTML]{FFFC9E}25.42} &
  \multicolumn{1}{c|}{\cellcolor[HTML]{FFFC9E}26.43} &
  \multicolumn{1}{c|}{\cellcolor[HTML]{FFFC9E}38.87} &
  \multicolumn{1}{c|}{\cellcolor[HTML]{FFFC9E}42.82} &
  \multicolumn{1}{c|}{\cellcolor[HTML]{FFFC9E}32.91} &
  \multicolumn{1}{c|}{\cellcolor[HTML]{FFFC9E}20.02} &
  \multicolumn{1}{c|}{\cellcolor[HTML]{FFFC9E}25.09} &
  \multicolumn{1}{c|}{\cellcolor[HTML]{FFFC9E}\textbf{23.26}} &
  \multicolumn{1}{c|}{\cellcolor[HTML]{FFFC9E}26.11} &
  \multicolumn{1}{c|}{\cellcolor[HTML]{FFFC9E}58.50} &
  \multicolumn{1}{c|}{\cellcolor[HTML]{FFFC9E}90.56} &
  \multicolumn{1}{c|}{\cellcolor[HTML]{FFFC9E}41.57} &
  \multicolumn{1}{c|}{\cellcolor[HTML]{FFFC9E}5.20} &
  \multicolumn{1}{c|}{\cellcolor[HTML]{FFFC9E}69.50} &
  \multicolumn{1}{c|}{\cellcolor[HTML]{FFFC9E}54.10} &
  38.69 \\ \hline
\rowcolor[HTML]{FFFC9E} 
H2O+KVCrush &
  \multicolumn{1}{c|}{\cellcolor[HTML]{FFFC9E}\textbf{25.48}} &
  \multicolumn{1}{c|}{\cellcolor[HTML]{FFFC9E}\textbf{27.01}} &
  \multicolumn{1}{c|}{\cellcolor[HTML]{FFFC9E}\textbf{39.02}} &
  \multicolumn{1}{c|}{\cellcolor[HTML]{FFFC9E}\textbf{43.20}} &
  \multicolumn{1}{c|}{\cellcolor[HTML]{FFFC9E}\textbf{33.23}} &
  \multicolumn{1}{c|}{\cellcolor[HTML]{FFFC9E}\textbf{20.78}} &
  \multicolumn{1}{c|}{\cellcolor[HTML]{FFFC9E}\textbf{25.67}} &
  \multicolumn{1}{c|}{\cellcolor[HTML]{FFFC9E}23.17} &
  \multicolumn{1}{c|}{\cellcolor[HTML]{FFFC9E}\textbf{26.33}} &
  \multicolumn{1}{c|}{\cellcolor[HTML]{FFFC9E}\textbf{62.00}} &
  \multicolumn{1}{c|}{\cellcolor[HTML]{FFFC9E}\textbf{90.48}} &
  \multicolumn{1}{c|}{\cellcolor[HTML]{FFFC9E}\textbf{41.87}} &
  \multicolumn{1}{c|}{\cellcolor[HTML]{FFFC9E}\textbf{5.23}} &
  \multicolumn{1}{c|}{\cellcolor[HTML]{FFFC9E}\textbf{69.50}} &
  \multicolumn{1}{c|}{\cellcolor[HTML]{FFFC9E}\textbf{54.32}} &
  \textbf{39.15} \\ \hline
\rowcolor[HTML]{CBCEFB} 
SnapKV &
  \multicolumn{1}{c|}{\cellcolor[HTML]{CBCEFB}{\color[HTML]{00B050} \textbf{25.70}}} &
  \multicolumn{1}{c|}{\cellcolor[HTML]{CBCEFB}{\color[HTML]{00B050} \textbf{29.96}}} &
  \multicolumn{1}{c|}{\cellcolor[HTML]{CBCEFB}38.93} &
  \multicolumn{1}{c|}{\cellcolor[HTML]{CBCEFB}43.90} &
  \multicolumn{1}{c|}{\cellcolor[HTML]{CBCEFB}35.05} &
  \multicolumn{1}{c|}{\cellcolor[HTML]{CBCEFB}20.44} &
  \multicolumn{1}{c|}{\cellcolor[HTML]{CBCEFB}{\color[HTML]{00B050} \textbf{26.89}}} &
  \multicolumn{1}{c|}{\cellcolor[HTML]{CBCEFB}\textbf{23.43}} &
  \multicolumn{1}{c|}{\cellcolor[HTML]{CBCEFB}26.17} &
  \multicolumn{1}{c|}{\cellcolor[HTML]{CBCEFB}74.00} &
  \multicolumn{1}{c|}{\cellcolor[HTML]{CBCEFB}\textbf{90.56}} &
  \multicolumn{1}{c|}{\cellcolor[HTML]{CBCEFB}41.96} &
  \multicolumn{1}{c|}{\cellcolor[HTML]{CBCEFB}5.54} &
  \multicolumn{1}{c|}{\cellcolor[HTML]{CBCEFB}69.25} &
  \multicolumn{1}{c|}{\cellcolor[HTML]{CBCEFB}{\color[HTML]{00B050} \textbf{56.16}}} &
  40.53 \\ \hline
\rowcolor[HTML]{CBCEFB} 
SnapKV+KVCrush &
  \multicolumn{1}{c|}{\cellcolor[HTML]{CBCEFB}25.62} &
  \multicolumn{1}{c|}{\cellcolor[HTML]{CBCEFB}29.48} &
  \multicolumn{1}{c|}{\cellcolor[HTML]{CBCEFB}\textbf{39.56}} &
  \multicolumn{1}{c|}{\cellcolor[HTML]{CBCEFB}\textbf{44.05}} &
  \multicolumn{1}{c|}{\cellcolor[HTML]{CBCEFB}\textbf{36.20}} &
  \multicolumn{1}{c|}{\cellcolor[HTML]{CBCEFB}\textbf{20.93}} &
  \multicolumn{1}{c|}{\cellcolor[HTML]{CBCEFB}26.24} &
  \multicolumn{1}{c|}{\cellcolor[HTML]{CBCEFB}23.16} &
  \multicolumn{1}{c|}{\cellcolor[HTML]{CBCEFB}\textbf{26.32}} &
  \multicolumn{1}{c|}{\cellcolor[HTML]{CBCEFB}\textbf{74.00}} &
  \multicolumn{1}{c|}{\cellcolor[HTML]{CBCEFB}90.54} &
  \multicolumn{1}{c|}{\cellcolor[HTML]{CBCEFB}\textbf{42.17}} &
  \multicolumn{1}{c|}{\cellcolor[HTML]{CBCEFB}\textbf{5.83}} &
  \multicolumn{1}{c|}{\cellcolor[HTML]{CBCEFB}\textbf{69.25}} &
  \multicolumn{1}{c|}{\cellcolor[HTML]{CBCEFB}55.48} &
  \textbf{40.59} \\ \hline
\rowcolor[HTML]{38FFF8} 
PyramidKV &
  \multicolumn{1}{c|}{\cellcolor[HTML]{38FFF8}25.53} &
  \multicolumn{1}{c|}{\cellcolor[HTML]{38FFF8}\textbf{29.89}} &
  \multicolumn{1}{c|}{\cellcolor[HTML]{38FFF8}38.67} &
  \multicolumn{1}{c|}{\cellcolor[HTML]{38FFF8}43.90} &
  \multicolumn{1}{c|}{\cellcolor[HTML]{38FFF8}35.04} &
  \multicolumn{1}{c|}{\cellcolor[HTML]{38FFF8}21.60} &
  \multicolumn{1}{c|}{\cellcolor[HTML]{38FFF8}\textbf{26.80}} &
  \multicolumn{1}{c|}{\cellcolor[HTML]{38FFF8}{\color[HTML]{00B050} \textbf{23.51}}} &
  \multicolumn{1}{c|}{\cellcolor[HTML]{38FFF8}26.37} &
  \multicolumn{1}{c|}{\cellcolor[HTML]{38FFF8}73.50} &
  \multicolumn{1}{c|}{\cellcolor[HTML]{38FFF8}90.56} &
  \multicolumn{1}{c|}{\cellcolor[HTML]{38FFF8}42.21} &
  \multicolumn{1}{c|}{\cellcolor[HTML]{38FFF8}5.08} &
  \multicolumn{1}{c|}{\cellcolor[HTML]{38FFF8}69.25} &
  \multicolumn{1}{c|}{\cellcolor[HTML]{38FFF8}55.36} &
  40.48 \\ \hline
\rowcolor[HTML]{38FFF8} 
PyramidKV+KVCrush &
  \multicolumn{1}{c|}{\cellcolor[HTML]{38FFF8}\textbf{25.57}} &
  \multicolumn{1}{c|}{\cellcolor[HTML]{38FFF8}29.48} &
  \multicolumn{1}{c|}{\cellcolor[HTML]{38FFF8}\textbf{38.97}} &
  \multicolumn{1}{c|}{\cellcolor[HTML]{38FFF8}\textbf{44.03}} &
  \multicolumn{1}{c|}{\cellcolor[HTML]{38FFF8}\textbf{35.75}} &
  \multicolumn{1}{c|}{\cellcolor[HTML]{38FFF8}\textbf{21.62}} &
  \multicolumn{1}{c|}{\cellcolor[HTML]{38FFF8}26.21} &
  \multicolumn{1}{c|}{\cellcolor[HTML]{38FFF8}23.28} &
  \multicolumn{1}{c|}{\cellcolor[HTML]{38FFF8}\textbf{26.52}} &
  \multicolumn{1}{c|}{\cellcolor[HTML]{38FFF8}\textbf{74.00}} &
  \multicolumn{1}{c|}{\cellcolor[HTML]{38FFF8}\textbf{90.56}} &
  \multicolumn{1}{c|}{\cellcolor[HTML]{38FFF8}\textbf{42.24}} &
  \multicolumn{1}{c|}{\cellcolor[HTML]{38FFF8}\textbf{5.17}} &
  \multicolumn{1}{c|}{\cellcolor[HTML]{38FFF8}\textbf{69.50}} &
  \multicolumn{1}{c|}{\cellcolor[HTML]{38FFF8}\textbf{55.49}} &
  \textbf{40.56} \\ \hline
KVCrush* &
  \multicolumn{1}{c|}{25.62} &
  \multicolumn{1}{c|}{29.48} &
  \multicolumn{1}{c|}{\cellcolor[HTML]{FFFFFF}{\color[HTML]{00B050} \textbf{39.56}}} &
  \multicolumn{1}{c|}{\cellcolor[HTML]{FFFFFF}{\color[HTML]{00B050} \textbf{44.05}}} &
  \multicolumn{1}{c|}{\cellcolor[HTML]{FFFFFF}{\color[HTML]{00B050} \textbf{36.20}}} &
  \multicolumn{1}{c|}{\cellcolor[HTML]{FFFFFF}{\color[HTML]{00B050} \textbf{21.62}}} &
  \multicolumn{1}{c|}{26.24} &
  \multicolumn{1}{c|}{23.28} &
  \multicolumn{1}{c|}{\cellcolor[HTML]{FFFFFF}{\color[HTML]{00B050} \textbf{26.52}}} &
  \multicolumn{1}{c|}{\cellcolor[HTML]{FFFFFF}{\color[HTML]{00B050} \textbf{74.00}}} &
  \multicolumn{1}{c|}{\cellcolor[HTML]{FFFFFF}{\color[HTML]{00B050} \textbf{90.56}}} &
  \multicolumn{1}{c|}{\cellcolor[HTML]{FFFFFF}{\color[HTML]{00B050} \textbf{42.24}}} &
  \multicolumn{1}{c|}{\cellcolor[HTML]{FFFFFF}{\color[HTML]{00B050} \textbf{5.83}}} &
  \multicolumn{1}{c|}{\cellcolor[HTML]{FFFFFF}{\color[HTML]{00B050} \textbf{69.50}}} &
  \multicolumn{1}{c|}{55.49} &
  {\color[HTML]{00B050} \textbf{40.68}} \\ \hline
\end{tabular}}
\caption{Performance comparison of KVCrush with PyramidKV, SnapKV and H2O on LongBench for LlaMa-3-8B-Instruct, Mistral-7B-Instruct-v0.2. Green bold indicates highest accuracy; similar background colors group each baseline method with its corresponding KVCrush result for easy comparison. KVCrush* (when paired with the top-performing KV compression method) achieves the highest accuracy across most datasets and the best average accuracy on both the \textit{LLaMa-3-8B-Instruct} and \textit{Mistral-7B-Instruct-v0.2} models.}
        \label{table:baseline_compare}
\end{table}
The accuracy of \emph{KVCrush}, in comparison with baseline methods, is detailed in Table~\ref{table:baseline_compare}. For this evaluation, a cache budget of 2048 was used. \emph{KVCrush} employs a static cache budget partitioning scheme, wherein $25\%$ of the total budget (i.e., 512) is allocated to \emph{KVCrush} for selecting representative tokens. The remaining $75\%$ is distributed to the accompanying method (H2O, SnapKV, or PyramidKV) for selecting high-attention tokens. The key insights are as follows:

\begin{itemize}
    \item \emph{KVCrush} when coupled with the best performing KV compression method (between H2O, SnapKV and PyramidKV) achieves the best accuracy for most of the datasets on both  \textit{LLaMa-3-8B-Instruct} and \textit{Mistral-7B-Instruct-v0.2} models.

    \item Using \textit{Mistral-7B-Instruct-v0.2}, it is the \textbf{only} method to achieve iso-accuracy  w.r.t. FullKV baseline on \textit{qmsum} and \textit{multi-news} datasets.

    \item \emph{KVCrush} achieves the best average accuracy on both \textit{LLaMa-3-8B-Instruct} and \textit{Mistral-7B-Instruct-v0.2} models.

\end{itemize}

\subsubsection{Inference Latency Overhead}
\begin{figure*}[t]
\floatconts
  {fig:latency_breakdown}
  {\caption{Latency breakdown on LongBench microbenchmark running on an Intel\textsuperscript{\textregistered} Xeon\textsuperscript{\textregistered} Platinum 8470 processor. H2O and H2O+KVCrush reduce KV cache size by $4\times$, which leads to a $3.2\times$ reduction in memory access latency. KVCrush adds only $\sim$0.2\% overhead while improving accuracy.}
}
  {
  \subfigure[Total Latency Breakdown]{
    \label{fig:latency_full}
    \begin{minipage}[t]{0.48\linewidth}
    \begin{tikzpicture}
    \begin{axis}[
        ybar stacked,
        bar width=14pt,
        width=\linewidth,
        height=5.5cm,
        enlargelimits=0.15,
        ylabel={Latency (normalized)},
        symbolic x coords={Full Cache, H2O, H2O+KVCrush},
        xtick=data,
        ymin=0, ymax=1.05,
        font=\small,
        legend style={
            at={(0.98,0.98)},
            anchor=north east,
            draw=none,
            fill=none,
            font=\small
        }
    ]
    \addplot+[ybar, fill=blue] coordinates {
        (Full Cache, 0.900)
        (H2O, 0.281)
        (H2O+KVCrush, 0.281)
    };
    \addplot+[ybar, fill=orange] coordinates {
        (Full Cache, 0.100)
        (H2O, 0.046)
        (H2O+KVCrush, 0.046)
    };
    \addplot+[ybar, fill=green] coordinates {
        (Full Cache, 0.000)
        (H2O, 0.003)
        (H2O+KVCrush, 0.005)
    };
    \legend{Memory, Compute, Pruning Overhead}
    \end{axis}
    \end{tikzpicture}
    \end{minipage}
  }\hfill
  \subfigure[Zoomed-in Pruning Overhead]{
    \label{fig:latency_zoom}
    \begin{minipage}[t]{0.48\linewidth}
    \begin{tikzpicture}
    \begin{axis}[
        ybar,
        bar width=14pt,
        width=\linewidth,
        height=5.5cm,
        ylabel={Pruning Overhead},
        symbolic x coords={H2O, H2O+KVCrush},
        xtick=data,
        ymin=0, ymax=0.01,
        font=\small
    ]
    \addplot+[ybar, fill=green] coordinates {
        (H2O, 0.003)
        (H2O+KVCrush, 0.005)
    };
    \end{axis}
    \end{tikzpicture}
    \end{minipage}
  }
  }
\end{figure*}
We evaluate the runtime latency of different KV cache strategies using a LongBench microbenchmark. 
The experiments were conducted on a system equipped with an Intel\textsuperscript{\textregistered} Xeon\textsuperscript{\textregistered} Platinum 8470 processor, simulating realistic CPU-bound server inference settings. The cache budget and associated compression parameters are detailed in Table~\ref{table:settings}.

Figure~\ref{fig:latency_breakdown} shows the normalized latency breakdown across three components: memory access, compute, and pruning overhead. We observe that while the full cache baseline incurs the highest memory access latency, both H2O and H2O+KVCrush significantly reduce this cost by reducing the KV cache size by $4\times$. Importantly, the additional pruning overhead introduced by KVCrush remains minimal—only $0.2\%$ higher than H2O—while achieving higher accuracy. This demonstrates that KVCrush can be deployed with negligible runtime cost while improving performance.

\subsection{Ablation Study}
\subsubsection{Selecting Anchor Points}
\label{sec:anchor_points} 
    \emph{KVCrush} employs an anchor point within the same binary space (a binary vector with a length equal to the number of heads) to form clusters of tokens based on their hamming distance to this anchor. As shown in Table~\ref{table:anchor}, \emph{KVCrush} outperforms the H2O baseline across all anchor-point strategies, including simple choices such as random, mean, and alternating 0–1 vectors. The 512-token budget shown is illustrative; similar trends are observed for other cache budgets.

\subsubsection{Cache budget partitioning}
\label{sec:cache_budget} 
    As discussed in Section~\ref{sec:kvcrush}, the cache budget $B$ allocated for the KV Cache is divided into two distinct segments: $B_{important}$, utilized by the baseline method to identify tokens with high attention weights, and $B_{representative}$, used by \emph{KVCrush} to select representative tokens. 
    
     Figure~\ref{fig:cache_budget_partitioning} illustrates the impact of varying the percentage of the budget allocated to \emph{KVCrush} on overall accuracy. For certain workloads, such as \textit{narrativeqa}, the optimal cache budget can reach up to $90\%$ of the total budget. In contrast, for other workloads, the optimal budget typically ranges between $20\%$ and $50\%$ of the available budget. In Section~\ref{sec:results_accuracy}, we compared our method with baseline approaches using a static cache budget partitioning scheme, allocating a fixed $25\%$ of the total budget to \emph{KVCrush}. This approach could potentially be enhanced by dynamically partitioning the budget based on attention weights distribution, which we plan to explore in future work. 
     
    \begin{figure*}[t]
\floatconts
  {fig:kmeans}
  {\caption{Accuracy-latency trade-off on GSM8K. KMeans offers slightly higher accuracy but with 200\% more latency, while KVCrush improves H2O with negligible cost. \textbf{Note:} \texttt{Phi3} and \texttt{LLaMA3} denote \texttt{Phi-3-mini-4k-instruct} and \texttt{Meta-Llama-3-8B-Instruct}.}}
  {
  \subfigure[Accuracy deficit w.r.t. Full-KV Baseline\label{fig:kmeans_acc}]{
    \begin{minipage}{0.45\linewidth}
    \centering
    \begin{tikzpicture}
    \begin{axis}[
        ybar,
        bar width=10pt,
        ymin=-18, ymax=0,
        ylabel={\shortstack{\textbf{Accuracy change (\%)}\\\textbf{w.r.t. Full-KV}}},
        symbolic x coords={Phi3, LLaMA3},
        xtick=data,
        enlarge x limits=0.2,
        width=\linewidth,
        height=5cm
    ]
    \addplot [fill=green!60!black] coordinates {(Phi3, -17.5) (LLaMA3, -1.5)};
    \addplot [fill=lime] coordinates {(Phi3, -10.0) (LLaMA3, -1.0)};
    \addplot [fill=cyan] coordinates {(Phi3, -7.5) (LLaMA3, -0.6)};
    \end{axis}
    \end{tikzpicture}
    \end{minipage}
  }
  \hfill
  \subfigure[Normalized inference Latency on Intel\textsuperscript{\textregistered} Xeon\textsuperscript{\textregistered} Platinum 8470\label{fig:kmeans_latency}]{
    \begin{minipage}{0.45\linewidth}
    \centering
    \begin{tikzpicture}
    \begin{axis}[
        ybar,
        bar width=10pt,
        ymin=0, ymax=3.5,
        ylabel={\shortstack{\textbf{Normalized Latency}\\\textbf{w.r.t. H2O}}},
        symbolic x coords={Phi3, LLaMA3},
        xtick=data,
        enlarge x limits=0.2,
        width=\linewidth,
        height=5cm
    ]
    \addplot [fill=green!60!black] coordinates {(Phi3, 1.0) (LLaMA3, 1.0)};
    \addplot [fill=lime] coordinates {(Phi3, 1.002) (LLaMA3, 1.003)};
    \addplot [fill=cyan] coordinates {(Phi3, 2.9) (LLaMA3, 3.1)};
    \end{axis}
    \end{tikzpicture}
    \end{minipage}
  }

  \vspace{0.5em}
  \centering
  \begin{tikzpicture}
  \begin{axis}[
      hide axis,
      xmin=0, xmax=1,
      ymin=0, ymax=1,
      legend style={at={(0.5,1.1)}, anchor=south, legend columns=3},
      legend image code/.code={
        \draw[#1, bar width=6pt, yshift=-0.1em] (0cm,0cm) rectangle +(0.3cm,0.3cm);
      }
  ]
  \addlegendimage{fill=green!60!black}
  \addlegendimage{fill=lime}
  \addlegendimage{fill=cyan}
  \legend{H2O, KVCrush, KMeans (100 iter)}
  \end{axis}
  \end{tikzpicture}
  }
\end{figure*}
    \begin{figure}[t]
\centering
\begin{tikzpicture}
\begin{axis}[
    width=0.49\textwidth,
    height=6cm,
    xlabel={\shortstack{\textbf{Percentage cache budget}\\\textbf{occupied by KVCrush}}},
    ylabel={\shortstack{\textbf{Accuracy Gain (\%)}\\\textbf{w.r.t. pure H2O}}},
    xmin=10, xmax=90,
    ymin=-1.5, ymax=5,
    xtick={10,20,...,90},
    ytick={-1,0,...,5},
    legend style={
    at={(1.02,0.5)},
    anchor=west,
    draw=none,
    font=\small
},
    legend cell align=left,
    grid=major,
    tick label style={font=\small},
    label style={font=\bfseries\small}
]

\addplot+[color=red, mark=*, thick] coordinates {
    (10,-1.2) (20,0.3) (30,0.5) (40,2.6) (50,3.2) (60,3.7) (70,4.5) (80,4.5) (90,4.5)
};
\addlegendentry{narrativeqa}

\addplot+[color=blue, mark=*, thick] coordinates {
    (10,0.3) (20,0.5) (30,1.9) (40,1.3) (50,1.3) (60,1.3) (70,1.5) (80,1.5) (90,0.6)
};
\addlegendentry{gov\_report}

\addplot+[color=orange!80!black, mark=*, thick] coordinates {
    (10,0.2) (20,0.4) (30,0.5) (40,0.5) (50,-0.1) (60,-0.6) (70,-0.6) (80,-0.6) (90,-0.6)
};
\addlegendentry{triviaqa}

\addplot+[color=green!60!black, mark=*, thick] coordinates {
    (10,0.3) (20,0.4) (30,0.4) (40,0.4) (50,0.8) (60,-0.1) (70,-0.1) (80,-0.1) (90,-0.1)
};
\addlegendentry{repobench\_p}

\addplot+[color=magenta, mark=none, thick, dashed] coordinates {
    (10,0.0) (20,0.0) (30,0.0) (40,0.0) (50,0.0) (60,0.0) (70,0.0) (80,0.0) (90,0.0)
};
\addlegendentry{Pure H2O}

\end{axis}
\end{tikzpicture}
\caption{Accuracy gain versus percentage of a fixed total cache budget allocated to KVCrush showing the trade-off between KVCrush and baseline allocation led to an empirical sweet spot of 20–50\% for most workloads, while a few (e.g., narrativeqa) benefit from higher allocations.}
\label{fig:cache_budget_partitioning}
\end{figure}
    \begin{figure*}[h]
\floatconts
  {fig:kvcrush_int}
  {\caption{Accuracy impact of integrating KVCrush with H2O, SnapKV, and PyramidKV on 2wikimqa. KVCrush improves both token and chunk-level pruning.}}
  {
  \subfigure[Token Level Eviction\label{fig:kvcrush_int_token}]{
    \begin{minipage}{0.45\linewidth}
    \centering
    \begin{tikzpicture}
    \begin{axis}[
        ybar,
        bar width=10pt,
        ymin=24, ymax=30,
        ylabel={\textbf{Accuracy - 2wikimqa}},
        symbolic x coords={H2O, SnapKV, PyramidKV},
        xtick=data,
        enlarge x limits=0.2,
        width=\linewidth,
        height=5cm,
        legend style={at={(0.5,-0.2)}, anchor=north, legend columns=2},
        legend image code/.code={
          \draw[#1, bar width=6pt, yshift=-0.1em] (0cm,0cm) rectangle +(0.3cm,0.3cm);
        }
    ]
    \addplot [fill=green!60!black] coordinates {(H2O, 26.0) (SnapKV, 27.5) (PyramidKV, 27.0)};
    \addplot [fill=cyan] coordinates {(H2O, 26.1) (SnapKV, 28.4) (PyramidKV, 28.4)};
    \legend{Without KVCrush, With KVCrush}
    \end{axis}
    \end{tikzpicture}
    \end{minipage}
  }
  \hfill
  \subfigure[Chunk (size: 8 tokens) Level Eviction\label{fig:kvcrush_int_chunk}]{
    \begin{minipage}{0.45\linewidth}
    \centering
    \begin{tikzpicture}
    \begin{axis}[
        ybar,
        bar width=10pt,
        ymin=24, ymax=30,
        ylabel={\textbf{Accuracy - 2wikimqa}},
        symbolic x coords={H2O, SnapKV, PyramidKV},
        xtick=data,
        enlarge x limits=0.2,
        width=\linewidth,
        height=5cm,
        legend style={at={(0.5,-0.2)}, anchor=north, legend columns=2},
        legend image code/.code={
          \draw[#1, bar width=6pt, yshift=-0.1em] (0cm,0cm) rectangle +(0.3cm,0.3cm);
        }
    ]
    \addplot [fill=green!60!black] coordinates {(H2O, 26.0) (SnapKV, 27.5) (PyramidKV, 27.0)};
    \addplot [fill=cyan] coordinates {(H2O, 26.9) (SnapKV, 27.8) (PyramidKV, 27.8)};
    \legend{Without KVCrush, With KVCrush}
    \end{axis}
    \end{tikzpicture}
    \end{minipage}
  }
  }
\end{figure*}
     
\subsubsection{KMeans vs KVCrush}
\label{sec:kmeans_comparison}    
    Traditional clustering algorithms exhibit inefficiencies in KV Cache compression due to two primary factors. Firstly, the large size of input vectors poses a challenge. For instance, in Llama3-8B model, each input key (or value) vector comprises 128 FP16 values. By employing \textit{kvcrush.representation}, a binary vector of length 32 (corresponding to the number of heads) is generated to represent each token (key-value pair). This approach not only reduces the data size by a factor of 64 but also enables faster distance computations using Hamming distances. Secondly, the time complexity for clustering algorithms is substantial. Utilizing \textit{kvcrush.grouping}, the selection of values to be retained in the cache is achieved in $O(2SD)$ time, in contrast to the $O(tS^2D)$ time complexity of k-means clustering.

    In Figure~\ref{fig:kmeans}, we compare \emph{KVCrush} and KMeans in terms of accuracy and inference latency using the GSM8K dataset. While KMeans achieves slightly higher accuracy than \emph{KVCrush} (Figure~\ref{fig:kmeans_acc}), it incurs approximately $200\%$ additional overhead in the inference latency pipeline. In contrast, \emph{KVCrush} provides a reasonable accuracy improvement over the pure H2O baseline, with an insignificant overhead of less than $0.5\%$.

 \subsubsection{Accuracy implications of integrating KVCrush with other methods}
\label{sec:kvcrush_int}
    We assess the impact on accuracy when integrating \emph{KVCrush} with other methods, utilizing cumulative attention-weights as the importance metric. Figure~\ref{fig:kvcrush_int} demonstrates the accuracy improvement provided by \emph{KVCrush} to the baseline methods for the \textit{2wikimqa} dataset. These baseline methods will prioritize selecting important tokens. As illustrated in Figure~\ref{fig:kvcrush_flow}, \emph{KVCrush} not only seamlessly integrates with these, but also improves upon their accuracies by selecting representative tokens (or pages) based on the head behavior. As shown in figure, \emph{KVCrush} also enhances the accuracy of the baseline methods even when they operate at chunks instead of individual tokens.
    
 \begin{figure*}[t]
\floatconts
  {fig:kvcrush_paged}
  {\caption{Accuracy difference w.r.t. full cache for Paged H2O and Paged-KVCrush on GSM8K and XSUM. Paged-KVCrush outperforms Paged-H2O across models.}}
  {
  \subfigure[GSM8K Accuracy using \texttt{Phi-3-mini-4k-instruct}]{
    \begin{minipage}{0.45\linewidth}
    \centering
    \begin{tikzpicture}
    \begin{axis}[
        ybar,
        bar width=10pt,
        ymin=-14, ymax=1,
        ylabel={\shortstack{\textbf{Accuracy change (\%)}\\\textbf{w.r.t. Full-KV}}},
        symbolic x coords={GSM8K-Flexible, GSM8K-Strict},
        xtick=data,
        enlarge x limits=0.2,
        width=\linewidth,
        height=5cm
    ]
    \addplot [fill=green!60!black] coordinates {(GSM8K-Flexible, -10.4) (GSM8K-Strict, -0.3)};
    \addplot [fill=cyan] coordinates {(GSM8K-Flexible, -5.1) (GSM8K-Strict, 0.9)};
    \end{axis}
    \end{tikzpicture}
    \end{minipage}
  }
  \hfill
  \subfigure[XSUM Accuracy using \texttt{Phi-3-mini-4k-instruct}]{
    \begin{minipage}{0.45\linewidth}
    \centering
    \begin{tikzpicture}
    \begin{axis}[
        ybar,
        bar width=10pt,
        ymin=-14, ymax=2,
        ylabel={\shortstack{\textbf{Accuracy change (\%)}\\\textbf{w.r.t. Full-KV}}},
        symbolic x coords={Rouge-1, Rouge-2, Rouge-L},
        xtick=data,
        enlarge x limits=0.2,
        width=\linewidth,
        height=5cm
    ]
    \addplot [fill=green!60!black] coordinates {(Rouge-1, -0.9) (Rouge-2, -13.5) (Rouge-L, -2.1)};
    \addplot [fill=cyan] coordinates {(Rouge-1, 1.6) (Rouge-2, -7.7) (Rouge-L, 0.1)};
    \end{axis}
    \end{tikzpicture}
    \end{minipage}
  }\par\medskip
  \subfigure[GSM8K Accuracy using \texttt{Meta-Llama-3-8B-Instruct}]{
    \begin{minipage}{0.45\linewidth}
    \centering
    \begin{tikzpicture}
    \begin{axis}[
        ybar,
        bar width=10pt,
        ymin=-3, ymax=1,
        ylabel={\shortstack{\textbf{Accuracy change (\%)}\\\textbf{w.r.t. Full-KV}}},
        symbolic x coords={GSM8K-Flexible, GSM8K-Strict},
        xtick=data,
        enlarge x limits=0.2,
        width=\linewidth,
        height=5cm
    ]
    \addplot [fill=green!60!black] coordinates {(GSM8K-Flexible, -2.5) (GSM8K-Strict, -2.4)};
    \addplot [fill=cyan] coordinates {(GSM8K-Flexible, -0.7) (GSM8K-Strict, 0.1)};
    \end{axis}
    \end{tikzpicture}
    \end{minipage}
  }
  \hfill
  \subfigure[XSUM Accuracy using \texttt{Meta-Llama-3-8B-Instruct}]{
    \begin{minipage}{0.45\linewidth}
    \centering
    \begin{tikzpicture}
    \begin{axis}[
        ybar,
        bar width=10pt,
        ymin=-1, ymax=4,
        ylabel={\shortstack{\textbf{Accuracy change (\%)}\\\textbf{w.r.t. Full-KV}}},
        symbolic x coords={Rouge-1, Rouge-2, Rouge-L},
        xtick=data,
        enlarge x limits=0.2,
        width=\linewidth,
        height=5cm
    ]
    \addplot [fill=green!60!black] coordinates {(Rouge-1, 0.8) (Rouge-2, 2.7) (Rouge-L, -0.1)};
    \addplot [fill=cyan] coordinates {(Rouge-1, 1.3) (Rouge-2, 3.5) (Rouge-L, 1.3)};
    \end{axis}
    \end{tikzpicture}
    \end{minipage}
  }

  \begin{center}
    \begin{tikzpicture}
      \begin{axis}[
        hide axis,
        xmin=0, xmax=1,
        ymin=0, ymax=1,
        legend style={at={(0.5,1.05)}, anchor=south, legend columns=2},
        legend image code/.code={
          \draw[#1, bar width=6pt, yshift=-0.1em] (0cm,0cm) rectangle +(0.3cm,0.3cm);
        }
      ]
      \addlegendimage{fill=green!60!black}
      \addlegendimage{fill=cyan}
      \addlegendentry{Paged H2O}
      \addlegendentry{Paged (H2O+KVCrush)}
      \end{axis}
    \end{tikzpicture}
  \end{center}
  }
\end{figure*}

\subsubsection{Evaluation of KVCrush in Paged KV settings}
\label{sec:kvcrush_paged} 
    To evaluate \emph{KVCrush} in paged KV settings, we utilized an H2O baseline that aggregates the row-wise sum of attention weights at the page level to evict low-importance pages. For \emph{KVCrush}, the binary vector of a page is formed by concatenating the binary head vectors of all tokens within that page. Figure~\ref{fig:kvcrush_paged} shows that Paged-\emph{KVCrush} outperforms paged-H2O on both GSM8K and XSUM datasets.

\section{Conclusion and Future Work}
\label{conclusion}
In this work, we presented an alternative approach to represent LLM tokens during inference. We showed that a compact representation paired with KVCrush compression algorithm leads to a substantial reduction in KV cache memory footprint. We demonstrated how we can use this method to achieve a memory-efficient LLM inference pipeline without compromising the quality of the generated tokens. In future work, we intend to investigate dynamic cache budget allocation and develop a more refined multi-anchoring approach.

\begin{spacing}{0.9}
\bibliography{acml23}

@article{gpt3,
  title={Language Models are Few-Shot Learners},
  author={Tom Brown and others},
  journal={Advances in Neural Information Processing Systems (NeurIPS)},
  year={2020}
}

@article{decomp,
  title={Get more with less: Synthesizing recurrence with kv cache compression for efficient llm inference.},
  author={Harry Dong and Xinyu Yang and Zhenyu Zhang and Zhangyang Wang and Yuejie Chi and Beidi Chen},
  journal={arXiv preprint arXiv:2402.09398},
  year={2024}
}

@article{gemini,
  title={Gemini 1.5: Unlocking multimodal understanding across millions of tokens of context},
  author={Petko Georgiev and others},
  journal={arXiv preprint arXiv:2403.05530},
  year={2024}
}

@article{vllm,
  title={Efficient Memory Management for Large Language Model Serving with PagedAttention},
  author={Woosuk Kwon and Zhuohan Li and Siyuan Zhuang and Ying Sheng and Lianmin Zheng and Cody Hao Yu and Joseph E. Gonzalez and Hao Zhang and Ion Stoica},
  journal={arXiv preprint arXiv:2309.06180},
  year={2023}
}

@article{mixed,
  title={Mixed-precision Quantization for Efficient LLM Deployment},
  author={Shiyao Li and Xuefei Ning  and Ke Hong and Tengxuan Liu and Luning Wang and Xiuhong Li and others},
  journal={International Conference on Neural Information Processing Systems, NeurIPS},
  year={2023}
}

@misc{llama,
  title = {{LLaMA-65B}: A 65-billion-parameter large language model},
  author = {Meta AI},
  year = {2023},
  url = {https://github.com/facebookresearch/llama},
}

@article{allyouneed,
  title={Attention is all you need},
  author={Ashish Vaswani and others},
  journal={International Conference on Neural Information Processing Systems, NeurIPS},
  year={2017}
}

@article{survey,
  title={Keep the Cost Down: A Review on Methods to Optimize {LLM' s KV-Cache} Consumption},
  author={Luohe Shi and Hongyi Zhang and Yao Yao and Zuchao Li and Hai Zhao},
  journal={arXiv preprint arXiv:2407.18003 },
  year={2024}
}

@article{h2o,
  title={{H2O}: Heavy-hitter oracle for efficient generative inference of large language models},
  author={Zhang, Zhenyu and Sheng, Ying and Zhou, Tianyi and Chen, Tianlong and Zheng, Lianmin and Cai, Ruisi and Song, Zhao and Tian, Yuandong and R{\'e}, Christopher and Barrett, Clark and others},
  journal={Advances in Neural Information Processing Systems},
  volume={36},
  year={2024}
}

@article{scissorhands,
  title={Scissorhands: Exploiting the persistence of importance hypothesis for llm kv cache compression at test time},
  author={Liu, Zichang and Desai, Aditya and Liao, Fangshuo and Wang, Weitao and Xie, Victor and Xu, Zhaozhuo and Kyrillidis, Anastasios and Shrivastava, Anshumali},
  journal={Advances in Neural Information Processing Systems},
  volume={36},
  year={2024}
}

@article{fastgen,
  title={Model tells you what to discard: Adaptive kv cache compression for llms},
  author={Ge, Suyu and Zhang, Yunan and Liu, Liyuan and Zhang, Minjia and Han, Jiawei and Gao, Jianfeng},
  journal={arXiv preprint arXiv:2310.01801},
  year={2023}
}

@article{snapkv,
  title={Snapkv: Llm knows what you are looking for before generation},
  author={Li, Yuhong and Huang, Yingbing and Yang, Bowen and Venkitesh, Bharat and Locatelli, Acyr and Ye, Hanchen and Cai, Tianle and Lewis, Patrick and Chen, Deming},
  journal={arXiv preprint arXiv:2404.14469},
  year={2024}
}

@article{pyramidkv,
  title={PyramidKV: Dynamic KV Cache Compression based on Pyramidal Information Funneling},
  author={Zhang, Yichi and Gao, Bofei and Liu, Tianyu and Lu, Keming and Xiong, Wayne and Dong, Yue and Chang, Baobao and Hu, Junjie and Xiao, Wen and others},
  journal={arXiv preprint arXiv:2406.02069},
  year={2024}
}

@article{mqa,
  title={Fast Transformer Decoding: One Write-Head is All You Need},
  author={Shazeer, Noam},
  journal={arXiv preprint arXiv:1911.02150},
  year={2019}
}

@article{gqa,
  title={GQA: Training Generalized Multi-Query Transformer Models from Multi-Head Checkpoints},
  author={Ainslie, Joshua and Lee-Thorp, James and de Jong, Michiel and Zemlyanskiy, Yury and Lebrón, Federico and Sanghai, Sumit},
  journal={arXiv preprint arXiv:2305.13245},
  year={2023}
}

@inproceedings{xiao2023smoothquant,
  title={SmoothQuant: Accurate and Efficient Post-Training Quantization for Large Language Models},
  author={Xiao, Guangxuan and Lin, Ji and Seznec, Mickael and Wu, Hao and Demouth, Julien and Han, Song},
  booktitle={Proceedings of the 40th International Conference on Machine Learning},
  pages={38087--38099},
  year={2023},
  organization={PMLR}
}

@inproceedings{liu2024llmqat,
  title={LLM-QAT: Data-Free Quantization Aware Training for Large Language Models},
  author={Liu, Zechun and Oguz, Barlas and Zhao, Changsheng and Chang, Ernie and Stock, Pierre and Mehdad, Yashar and Shi, Yangyang and Krishnamoorthi, Raghuraman and Chandra, Vikas},
  booktitle={Findings of the Association for Computational Linguistics: ACL 2024},
  pages={467--484},
  year={2024},
  address={Bangkok, Thailand and virtual meeting},
  publisher={Association for Computational Linguistics}
}

@inproceedings{sheng2023flexgen,
  title={FlexGen: High-Throughput Generative Inference of Large Language Models with a Single GPU},
  author={Sheng, Ying and others},
  booktitle={Proceedings of the 40th International Conference on Machine Learning},
  pages={31094--31116},
  year={2023},
  organization={PMLR}
}

@inproceedings{mq1,
  title={HAQ: Hardware-Aware Automated Quantization with Mixed Precision},
  author={Wang, Kuan and Liu, Zhijian and Lin, Yujun and Lin, Ji and Han, Song},
  booktitle={Proceedings of the IEEE Conference on Computer Vision and Pattern Recognition (CVPR)},
  pages={},
  year={2019}
}

@inproceedings{mq2,
  title={HAWQ-V3: Dyadic Neural Network Quantization},
  author={Yao, Zhewei and Dong, Zhen and Zheng, Zhangcheng and Gholami, Amir and Yu, Jiali and Tan, Eric and Wang, Leyuan and Huang, Qijing and Wang, Yida and Mahoney, Michael W. and Keutzer, Kurt},
  booktitle={38th International Conference on Machine Learning (ICML)},
  pages={},
  year={2021},
  organization={}
}

@inproceedings{mq3,
  title={Post Training Mixed Precision Quantization of Neural Networks Using First-Order Information},
  author={Chauhan, Arun and Tiwari, Utsav and Vikram, N R},
  booktitle={Proceedings of the IEEE/CVF International Conference on Computer Vision (ICCV) Workshops},
  pages={1343--1352},
  year={2023}
}

@article{mikv,
  title={No Token Left Behind: Reliable KV Cache Compression via Importance-Aware Mixed Precision Quantization},
  author={Yang, June Yong and Kim, Byeongwook and Bae, Jeongin and Kwon, Beomseok and Park, Gunho and Yang, Eunho and Kwon, Se Jung and Lee, Dongsoo},
  journal={arXiv preprint arXiv:2402.18096},
  year={2024}
}

@inproceedings{memrec,
  title = {{Mem-Rec}: {M}emory Efficient Recommendation System using Alternative Representation},
  author = {Jha, Gopi Krishna and Thomas, Anthony and Jain, Nilesh and Gobriel, Sameh and Rosing, Tajana and Iyer, Ravi},
  booktitle = {Proceedings of the 15th Asian Conference on Machine Learning},
  pages = {518--533},
  year = {2024},
  editor = {},
  volume = {},
  series = {},
  month = {},
  publisher = {},
  url = {}
}

@article{gholami2024ai,
  title={AI and Memory Wall},
  author={Gholami, Amir and Yao, Zhewei and Kim, Sehoon and Hooper, Coleman and Mahoney, Michael W. and Keutzer, Kurt},
  journal={IEEE Micro},
  year={2024},
  volume={44},
  number={2},
  pages={},
  doi={}
}

@inproceedings{sun2019bert4rec,
  title={BERT4Rec: Sequential recommendation with bidirectional encoder representations from transformer},
  author={Sun, Fei and Liu, Jun and Wu, Jian and Pei, Changhua and Lin, Xiao and Ou, Wenwu and Jiang, Peng},
  booktitle={Proceedings of the 28th ACM international conference on information and knowledge management},
  pages={1441--1450},
  year={2019}
}

@article{dosovitskiy2020image,
  title={An image is worth 16x16 words: Transformers for image recognition at scale},
  author={Dosovitskiy, Alexey},
  journal={arXiv preprint arXiv:2010.11929},
  year={2020}
}

@article{raffel2020exploring,
  title={Exploring the limits of transfer learning with a unified text-to-text transformer},
  author={Raffel, Colin and Shazeer, Noam and Roberts, Adam and Lee, Katherine and Narang, Sharan and Matena, Michael and Zhou, Yanqi and Li, Wei and Liu, Peter J},
  journal={Journal of machine learning research},
  volume={},
  number={},
  pages={},
  year={2020}
}

@inproceedings{isaev2023scaling,
  title={Scaling Infrastructure to Support Multi-Trillion Parameter LLM Training},
  author={Isaev, Mikhail and McDonald, Nic and Vuduc, Richard},
  booktitle={Proceedings of the 50th Annual International Symposium on Computer Architecture (ISCA)},
  year={2023},
  organization={IEEE}
}

@article{streamingllm,
  title={Efficient streaming language models with attention sinks},
  author={Xiao, Guangxuan and Tian, Yuandong and Chen, Beidi and Han, Song and Lewis, Mike},
  journal={arXiv preprint arXiv:2309.17453},
  year={2023}
}

@article{longbench,
      title={LongBench: A Bilingual, Multitask Benchmark for Long Context Understanding}, 
      author={Yushi Bai and Xin Lv and Jiajie Zhang and Hongchang Lyu and Jiankai Tang and Zhidian Huang and Zhengxiao Du and Xiao Liu and Aohan Zeng and Lei Hou and Yuxiao Dong and Jie Tang and Juanzi Li},
      year={2024},
      eprint={2308.14508},
      archivePrefix={arXiv},
      primaryClass={cs.CL},
      journal={Xiv preprint arXiv:2308.14508}, 
}

@article{dynamickv,
      title={DynamicKV: Task-Aware Adaptive KV Cache Compression for Long Context LLMs}, 
      author={Xiabin Zhou and Wenbin Wang and Minyan Zeng and Jiaxian Guo and Xuebo Liu and Li Shen and Min Zhang and Liang Ding},
      year={2025},
      eprint={2412.14838},
      archivePrefix={arXiv},
      primaryClass={cs.CL},
    journal={arXiv preprint arXiv:2412.14838},
      url={}, 
}

@article{lacache,
      title={LaCache: Ladder-Shaped KV Caching for Efficient Long-Context Modeling of Large Language Models}, 
      author={Dachuan Shi and Yonggan Fu and Xiangchi Yuan and Zhongzhi Yu and Haoran You and Sixu Li and Xin Dong and Jan Kautz and Pavlo Molchanov and Yingyan and Lin},
      year={2025},
      eprint={2507.14204},
      archivePrefix={arXiv},
      primaryClass={cs.LG},
    journal={arXiv preprint arXiv:2507.14204},
      url={}, 
}
\end{spacing}

\end{document}